# Robust non-local means filter for ultrasound image denoising

**Hamid Reza Shahdoosti**[1,*]

[1] Department of Electrical Engineering, Hamedan University of Technology, Hamedan, Iran
email address: h.doosti@hut.ac.ir

## ABSTRACT

This paper introduces a new approach to non-local means image denoising. Instead of using all pixels located in the search window for estimating the value of a pixel, we identify the highly corrupted pixels and assign less weight to these pixels. This method is called robust non-local means.

Numerical and subjective evaluations using ultrasound images show good performances of the proposed denoising method in recovering the shape of edges and important detailed components, in comparison to traditional ultrasound image denoising methods.

## 1. INTRODUCTION

Owing to the fact that ultrasound images are noninvasive and harmless to the human body, they are widely used in different medical applications. These images, however, are corrupted by speckle noise due to the imaging principle. The speckle noise can reduce the contrast of these images and blur the edges and details, such that the reliability of the image decreases and the clinical diagnosis is affected. One can make use of image processing methods for suppressing speckle noise to improve the quality and enhance the diagnostic potential of ultrasound image.

Different methods have been proposed during the last decades for the speckle reduction of ultrasound images. By transforming the multiplicative noise of the ultrasound images into an additive Gaussian one in Log compressed images, the NCD (Nonlinear Coherent Diffusion) method [1] was introduced. The developed version of the Perona and Malik diffusion method called SRAD (Speckle Reducing Anisotropic Diffusion) which casts the typical spatial adaptive filers into diffusion model was proposed to remove speckle noise of ultrasound images [2,3]. Due to the fact that the diffusion coefficient depends on noise in each iteration step, OSRAD (Oriented SRAD) filter [4] was proposed which is an extension of SRAD. This method was motivated by analyzing the properties of the numerical framework associated with SRAD filter by means of a semi-explicit scheme. This method is based on matrix anisotropic diffusion and is able to make the several diffusions across to the curvature directions. Several methods based on Total Variation (TV) minimization approach for ultrasound image despeckling have been introduced (for example see [5,6]). These methods use the norm of TV as the smoothing term. In addition, some methods use Rayleigh distribution to model the distribution of noise in the ultrasound images [7,8]. Based on the Rayleigh distribution modeling, different ultrasound image denoising methods were proposed [9–11]. The TV norm exploited in these methods for preservation of edges is not in harmony with speckle removing due to the fact that the noise is very large. In other words, the details are blurred when noise is suppressed or details are preserved but the noise is not well eliminated.

Morphological component analysis using nonlinear operators (erosion, dilation, opening and closing), is one of the effective tools for decomposition of the given signal or image into superposed contributions from different sources. Owing to the fact that the noise free signal or image and the noise are fundamentally different morphological components, one can easily consider them as two different sources and adopt the morphological component analysis to separate the unknown noise from the noiseless signal or image. Methods based on morphological analysis can easily handle the denoising task when the image is contaminated by the non-additive noise such as salt and pepper noise and multiplicative noise [12-13].

Denoising methods applied to ordinary images [14-17] can be used for ultrasound image despeckling after applying a Log transformation to them. One of the most powerful denoising methods is non-local means which can be used along with a Log transformation to reduce noise of ultrasound images. In this method the similarity between two pixels depends on the similarity of the patches surrounding the pixels [18]. A new modification to the weights of the NL-means (non-local means) method is proposed in this paper such that the more a pixel is noisy, the less weight it has.

We call this method robust NL-means. Experimental results demonstrate that this method has a better performance when denoising ultrasound images corrupted by multiplicative noise compared with the traditional NL-means method.

The organization of this paper is as follows. In the next section, the NL-means algorithm is briefly reviewed and the proposed method is presented in detail. Experimental results, including a comparison with the traditional NL-means methods, are given in section 3. Finally, section 4 presents the conclusions.

## 2. PROPOSED ROBUST NL-MEANS ALGORITHM

Suppose that a discrete noisy image is given denoted by $v = \{v(i) | i \in I\}$, the estimated value for the ith pixel denoted by $NL(i)$, is computed as a weighted average of all the gray levels in the image

$$NL(i) = \sum_{j \in I} w(i,j) v(j) \quad (1)$$

in which $w(i,j)$ is the weight corresponding to the value $v(j)$ for estimating the ith pixel. Although in the traditional definition of the NL-means algorithm, the pixels of the whole image are considered in the estimation of the intensity of each pixel, for practical applications, the number of pixels considered in the weighted average is limited to a neighborhood search window $S_i$ around the ith pixel. In more detail, the weight $w(i,j)$ is determined by the similarity between the intensities of the local patches $v(N_i)$ and $v(N_j)$ centered around the ith pixel and jth pixel, respectively, such that $0 \leq w(i,j) \leq 1$ and $\sum_j w(i,j) = 1$, where $N_l$ denotes a square neighborhood, centered around the $l$th pixel and is surrounded by the search window $S_i$. For measuring the similarity, a decreasing function (usually an exponential function) of the weighted Euclidean distance $\|v(N_i) - v(N_j)\|_{2,s}^2$ is used, where $\|v(N_i) - v(N_j)\|_{2,s}^2$ is the traditional L$_2$-norm which is convolved with a Gaussian kernel of standard deviation $s$. Owing to the fact that closer pixels are more dependent, they should have larger weights in the window comparison. Therefore, the Gaussian kernel is utilized to assign different weights corresponding to the pixel position, i.e., the central pixel in the window contributes more to the distance compared with the peripheral pixels. The weights $w(i,j)$ are defined as follows:

$$w(i,j) = \frac{1}{C(i)} e^{-\frac{\|v(N_i) - v(N_j)\|_{2,s}^2}{h^2}} \quad (2)$$

in which $C(i)$ is the normalizing constant:

$$C(i) = \sum_{j \in S_i} e^{-\frac{\|v(N_i) - v(N_j)\|_{2,s}^2}{h^2}} \quad (3)$$

which guarantees that $\sum_j w(i,j) = 1$, and $h$ is the function parameter by which the decay of the weights as a function of the Euclidean distances is controlled. Because of the fast decay of the exponential function, the computed weights for large Euclidean distances are approximately zero, i.e., this function acts as an automatic threshold. Therefore, the NL-means compares both the intensity level in a single point and the geometrical configuration in a whole neighborhood to achieve a more robust comparison compared with the neighborhood filters.

By using the Euclidean distance for the noisy neighborhoods, one can conclude [18]:

$$E \|v(N_i) - v(N_j)\|_{2,s}^2 = \|u(N_i) - u(N_j)\|_{2,s}^2 + 2 s_n^2 \quad (4)$$

in which $u$ is the noise free image, $v$ denotes the noisy image and $s_n$ is the standard deviation of noise. Due to the fact that the Euclidean distance preserves the order of the similarity between pixels in the expected value sense, the algorithm is robust.

The ability of the NL-means method to remove the noise and preserve the edges, has made this algorithm one of the most successful ones for image denoising. However, when the noise increases, the performance of this method decreases significantly and the denoised image suffers from blurring and lack of image details [19]. The main reason is that all pixels including highly corrupted pixels and non-corrupted pixels, located in the search window $S_i$, contribute to estimate the value of the ith pixel. Note that less weights should be given to the highly corrupted pixels by an ideal algorithm when the value of the ith pixel is estimated. Therefore, the algorithm should identify the highly corrupted pixels and assign less weights to these pixels. For this purpose, we introduce a new modification to the weights of the NL-means algorithm. In our strategy, a Gaussian filter is firstly applied to the noisy image to reduce the effect of noise as much as possible. Then, to measure the level of corruption for each pixel, an approach based on the local difference value is used. Finally, a decreasing function of this value is taken into account in the calculation of the weights of the NL-means algorithm.

Given a noisy image $v(j)$, the proposed robust NL-means method, filters the image by a Gaussian kernel to obtain the denoised image $\hat{v}(j)$. If the difference value between $v(j)$ and $\hat{v}(j)$ is large, the jth pixel is considered as a corrupted pixel. So the proposed algorithm uses the denoised image $\hat{v}(j)$ to obtain the modified weights as follows:

$$NL(i) = \sum_{j \in S_i} w(i,j) v(j) \tag{5}$$

$$w(i,j) = \frac{1}{C(i)} e^{-\frac{\|v(N_i)-v(N_j)\|_{2,s}^2}{h_1^2}} e^{-\frac{|v(j)-\hat{v}(j)|}{h_2}} \tag{6}$$

where $h_1$ and $h_2$ are the function parameters controlling the decay of the weights as a function of the Euclidean distances and that of the weights as a function of the level of noise of the jth pixel, respectively. In addition, $C(i)$ is the normalizing constant. As described before, due to the fast decay of the exponential function, it can be used as an automatic threshold such that if the difference value between $v(j)$ and $\hat{v}(j)$ is larger than the threshold, the jth pixel is a highly corrupted pixel and the weight of it in estimating the ith pixel is approximately set to zero.

## 3. NUMERICAL EXPERIMENTS

The method proposed in this paper is used to remove the speckle noise of medical ultrasound images. The proposed method is compared with several speckle suppression methods such as the Frost filter [20], LEE method [21], SRAD method [2], and ML-means method [18]. The parameters of the proposed method are set to: the size of the search window $S_i = 21 \times 21$, the size of the neighborhood window = $7 \times 7$, the value of the parameter $h_1 = 9 s_n$, and the values of the parameter $h_2 = 148/s_n$, where $s_n$ is the standard deviation of noise.

Here, two real ultrasound images are used in the experiments which are shown in Figs. 1(a) and 2(a). The results for the Frost filter are shown in Figs. 1(b) and 2(b), those for the LEE method are shown in Figs. 1(c) and 2(c), those for the SRAD method are shown in Figs. 1(d) and 2(d), those for the SRAD method are shown in Figs. 1(e) and 2(e), and those for the proposed method are shown in Figs. 1(f) and 2(f).

The human eye may not be merely sufficient for assessment of the despeckling capability as there are tiny details which are not distinguishable by the human perception. Another problem in assessing real ultrasound images is that original noise-free images is not available. Hence, it is not possible to employ traditional criteria such as PSNR (Peak Signal to Noise Ratio), EPI (Edge Preservation Index), and SSIM (Structural Similarity) [22-24]. However, it is still needed to observe the speckle suppression, edge preservation, and presence of artifacts after the despeckling process. Therefore, an evaluation method that is different from those used for ordinary images has to be designed for ultrasound images. Here, the NIQE (No Reference Image Quality Assessment) [25] is used which does not need the original noise free image to evaluate the performance of despeckling algorithms. The lower the value of NIQE, the better the quality of the denoised image. The NIQE results for the two ultrasound images are shown in Tables 1 and 2, respectively.

As can be seen from the figures, the proposed method not only removes the noise, but also maintains details better than other methods. In addition, the proposed method performs better than other despackling methods in terms of the NIQE index.

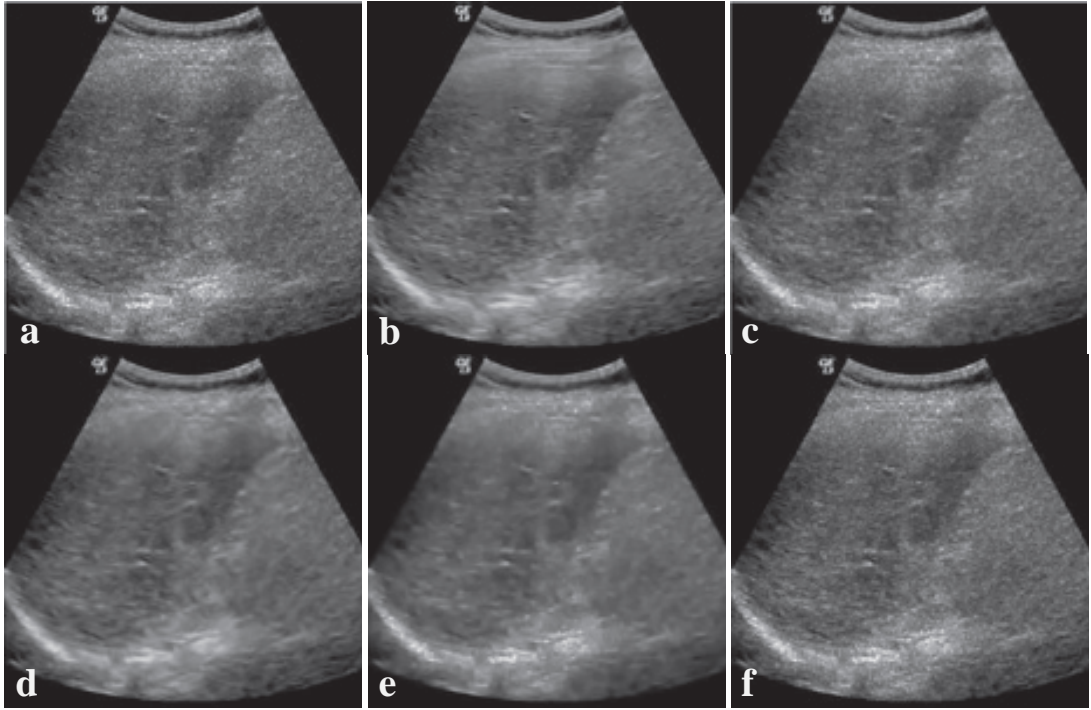

**Fig. 1**. Ultrasound image denoising using different methods. (a) Noisy image. (b) Denoised image using FROST [20]. (c) Denoised image using LEE [21]. (d) Denoised image using SRAD [2]. (e) Denoised image using NL-means [18]. (f) Denoised image using robust NL-means.

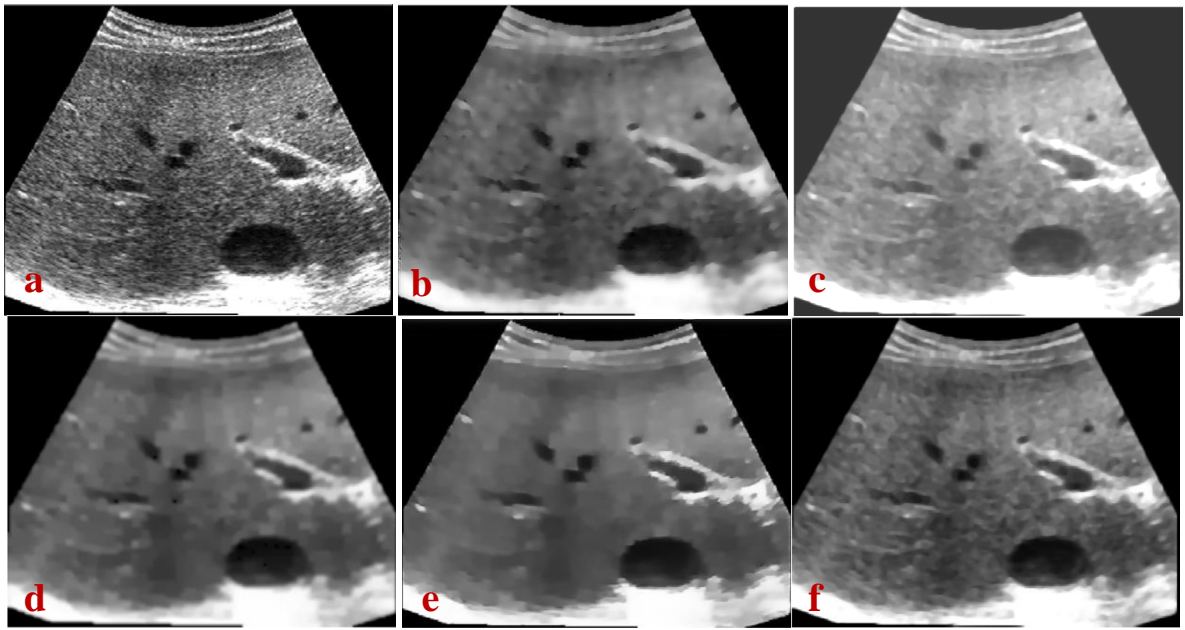

**Fig. 2**. Ultrasound image denoising using different methods. (a) Noisy image. (b) Denoised image using FROST [20]. (c) Denoised image using LEE [21]. (d) Denoised image using SRAD [2]. (e) Denoised image using NL-means [18]. (f) Denoised image using robust NL-means.

Table 1. NIQE values for the first ultrasound images.

| Method | NIQE |
|---|---|
| **FROST method** | 27.14 |
| **LEE method** | 28.37 |
| **SRAD method** | 28.70 |
| **NL-means method** | 27.57 |
| **Proposed method** | 25.41 |

Table 2. NIQE values for the second ultrasound images.

| Method | NIQE |
|---|---|
| **FROST method** | 25.28 |
| **LEE method** | 26.71 |
| **SRAD method** | 26.45 |
| **NL-means method** | 25.96 |
| **Proposed method** | 23.32 |

### 4. CONCLUSION

This paper has presented a new despeckling method for removing noise of ultrasound images. The proposed method is a robust one in which the highly corrupted pixels were identified and less weights were assigned to these pixels. The proposed algorithm were tested on two real ultrasound images. The result of the experiments are shown in the figures. In addition, the NIQE metric was used to evaluate the results of various ultrasound despeckling methods. As it was shown in the figures and tables, the proposed method can better remove noise of real ultrasound images.